\documentclass[conference]{IEEEtran}
\usepackage{cite}

\ifCLASSINFOpdf
   \usepackage[pdftex]{graphicx}
\else
   \usepackage[dvips]{graphicx}
\fi
\usepackage[cmex10]{amsmath}
\usepackage{multirow}

\begin{document}
%
\title{Novelty-Organizing Team of Classifiers in Noisy and Dynamic Environments}

\author{\IEEEauthorblockN{Danilo Vasconcellos Vargas}
\IEEEauthorblockA{Graduate School of Information\\ Science and Electrical Engineering\\
Kyushu University\\
Fukuoka, Japan\\
Email: vargas@cig.ees.kyushu-u.ac.jp}
\and
\IEEEauthorblockN{Hirotaka Takano}
\IEEEauthorblockA{Faculty of Information Science\\ and Electrical Engineering\\
Kyushu University\\
Fukuoka, Japan\\
Email: takano@cig.ees.kyushu-u.ac.jp}
\and
\IEEEauthorblockN{Junichi Murata}
\IEEEauthorblockA{Faculty of Information Science\\ and Electrical Engineering\\
Kyushu University\\
Fukuoka, Japan\\
Email: murata@cig.ees.kyushu-u.ac.jp}}


%


\IEEEoverridecommandlockouts
\IEEEpubid{\makebox[\columnwidth]{978-1-4799-7492-4/15/\$31.00~
		\copyright2015
IEEE \hfill} \hspace{\columnsep}\makebox[\columnwidth]{ }}

\maketitle

\begin{abstract}
In the real world, the environment is constantly changing with the input variables under the effect of noise.
However, few algorithms were shown to be able to work under those circumstances.
Here, Novelty-Organizing Team of Classifiers (NOTC) is applied to the continuous action mountain car as well as two variations of it: a noisy mountain car and an unstable weather mountain car.
These problems take respectively noise and change of problem dynamics into account.
Moreover, NOTC is compared with NeuroEvolution of Augmenting Topologies (NEAT) in these problems, revealing a trade-off between the approaches.
While NOTC achieves the best performance in all of the problems, NEAT needs less trials to converge.
It is demonstrated that NOTC achieves better performance because of its division of the input space (creating easier problems).
Unfortunately, this division of input space also requires a bit of time to bootstrap.
\end{abstract}


\section{Introduction}

Everything in the real world is naturally dynamic and noisy.
Although most of the systems employ some type of pre-processing to treat these type of problems, sometimes the pre-processing systems may face some unexpected new dynamics which they were not prepared for or the noise may suddenly change in type and amplitude.
Actually, learning algorithms seems the most natural solution to these problems, since they were developed right from the beginning with the idea of adaptation.
That is, all of them are in principle capable of learning new dynamics or noise variations of the problem on the fly.

In this paper, NOTC and NEAT are tested on problems that have noise and need a certain degree of adaptability.
On one hand, NEAT is a promising direct encoding topology evolving neuroevolution algorithm with a complexification philosophy (the chromosome starts simple and gets complex over time) \cite{stanley2002evolving}.
On the other hand, NOTC is a learning classifier system based algorithm with a divide and conquer philosophy, it currently evolves a fixed topology neural network with a direct encoding genome.
NOTC has the following distinct features:
\begin{itemize}
	\item Novelty Map Population - Experiments show that this type of population allows for better adaptation at the same time that it is not sensitive to noise;
	\item Dual (team-individual) Fitness - The dual fitness presents a way to join Michigan and Pittsburgh approaches, leveraging the benefits from both points of view;
	\item Hall of Fame - With the Hall of Fame it is possible to keep the best combination of individuals. This is important to join Michigan and Pittsburgh approaches.
\end{itemize}		

NOTC was first proposed and superficially described in \cite{vargas2014novelty}, with applications only to pole balancing and a discrete version of mountain car. 
Here we explain NOTC more deeply and apply it to continuous action mountain car, noisy mountain car and unstable weather mountain car (a problem requiring some level of adaptability). 
Moreover, a comparison between NOTC and NEAT is done in all of the problems, revealing a trade-off between the two approaches.

It is verified that NOTC achieves the best performance in all of the problems because of its ability to divide the input space creating smaller easier problems.
NEAT, on the other hand, needs less trials to converge because it does not need to divide the input space as well as it only has one problem to solve.

\begin{figure*}[!th]
\centering
\includegraphics[scale=1.0]{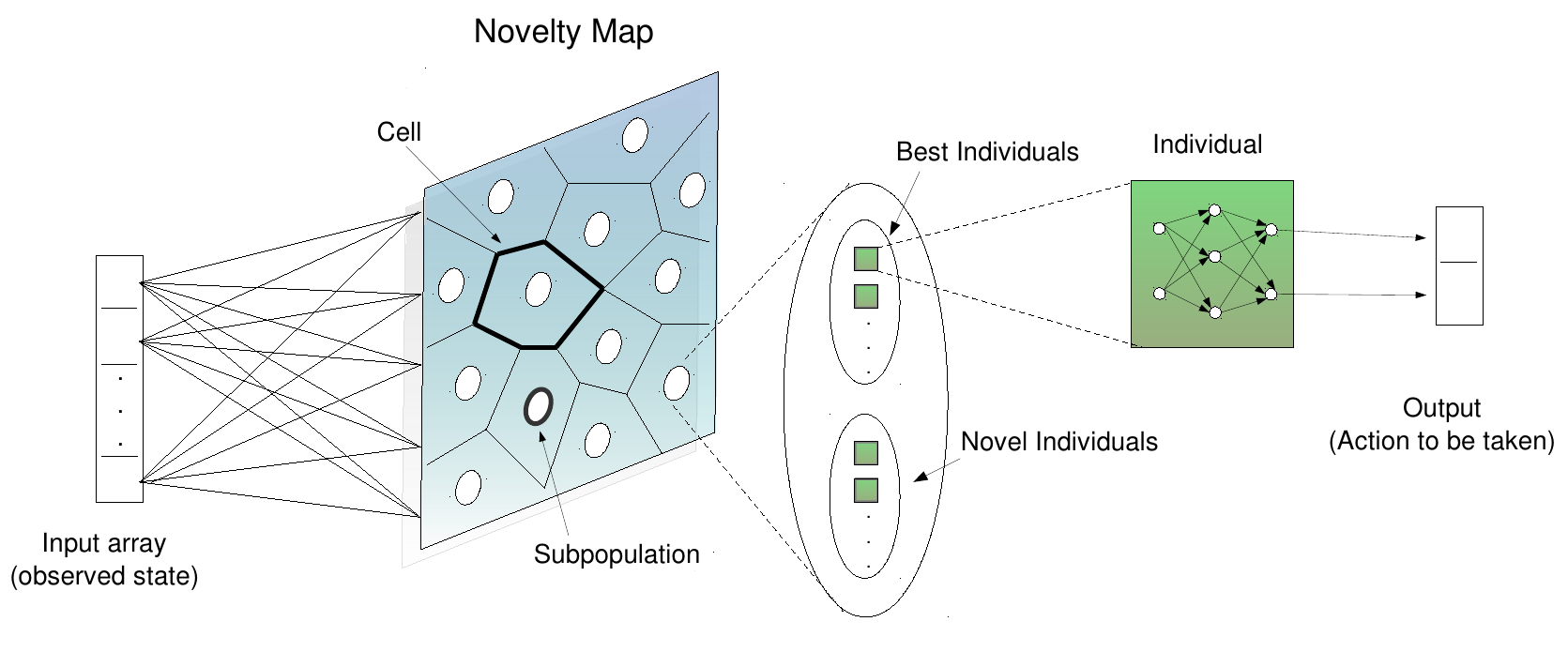}
\caption{NOTC's structure is illustrated. Notice that this is just an example, the size of the Novelty Map, number of inputs and so on were not taken into consideration. }
\label{structure}
\end{figure*}

\section{Related Work}



Allow us to divide the literature in two lines of thought:
\begin{itemize}
	\item Divide and Conquer Approach - division of the problem into easier ones and the use of simple computational models to solve those easier problems;
	\item Complexification Approach - start with simple solutions, while increasing the complexity of solutions over time.
\end{itemize}
On one hand, Learning Classifier Systems (LCS) falls within the divide and conquer approach, where a set of agents with condition-action-prediction rules cooperate and compete in an evolutionary system to solve the problem at hand \cite{holland1977cognitive,holmes2002learning}.
The condition coded by each agent automatically divides the input space, creating smaller problems, therefore the coded solutions does not need to be complex.
On the other hand, many evolutionary algorithms using variable length genomes fall within the complexification approach. 
For example, algorithms evolving both the topology and parameters of neural networks complexify the network over time \cite{floreano2008neuroevolution}.

In the following, there is a brief review of the LCS's and Neuroevolution's literature's related with this article.
Here, we will restrain the review to only continuous action LCS algorithms and some of most salient Neuroevolution methods.
For a complete review of both LCS's and Neuroevolution's literature please refer to \cite{urbanowicz2009learning,lanzi2000roadmap} and \cite{floreano2008neuroevolution}, respectively.

LCSs with continuous actions were applied to function approximation first with the XCSF algorithm \cite{wilson2002classifiers, butz2008function,tran2007xcsf}, later with variants using fuzzy logic  \cite{valenzuela1991fuzzy, bull2002accuracy, casillas2007fuzzy}, neural-based LCS algorithms \cite{bull2002using, bull2002accuracy} and genetic programming-based algorithms \cite{iqbal2012xcsr}.
Regarding reinforcement learning problems, LCSs with discrete actions were used to solve complex mazes \cite{lanzi2005xcs}, cart-pole balancing \cite{Twardowski1993a,bonarini1996evolutionary} and the two-actions mountain car \cite{lanzi2006classifier} problems.
Continuous action LCSs were applied to control robotic arms  \cite{stalph2012learning,butz2008context}, navigation problems \cite{bonarini2000fuzzy,howard2009towards}, complex mazes \cite{vargas2013self,vargas2013aself} and dynamical mazes \cite{vargas2013continuous}.
NOTC, specifically, was applied to pole balancing and a discrete action of mountain car in \cite{vargas2014novelty}.
NOTC related algorithm without the concepts of team and dual fitness, Novelty-Organizing Classifiers (NOC), was applied to continuous mazes and classification problems \cite{vargas2014noc}.

Neuroevolution, where both the structure and parameters are evolved, is a relatively new research area.
Therefore, there are fewer algorithms.
To cite some: GNARL \cite{angeline1994evolutionary}, NEAT \cite{stanley2002evolving}, EANT \cite{kassahun2005efficient} and EPNet \cite{yao1997new}.

\section{NOTC's Structure}

In divide and conquer approaches, an algorithm has usually two distinct procedures, one for breaking the problem (divide procedure) and another to build the solution for the problem pieces (conquer procedure).
NOTC uses novelty map as the divide procedure and multilayer perceptron as the conquer procedure.

The details of NOTC's structure is shown in Figure~\ref{structure}.
Basically, its components are:
\begin{itemize}
\item A Novelty Map population;
\item Subpopulations divided in two groups (best and novel);
\item Individuals.
\end{itemize}

\subsection{Subpopulation}
\label{subpopulation}

The subpopulation is a set of individuals. 
It is divided in two groups (best and novel).
The best and novel groups are useful to increase the diversity of the population and attain a good balance between exploration and exploitation.
In one hand, individuals inside the best group are the best individuals, in the sense that they were already tested before (survived the last generation). 
On the other hand, individuals from the novel group were recently created (created in the last generation).


\subsection{Individuals}

Individuals can be any computational model. 
The individuals used in this paper are feedforward neural networks with a single hidden layer containing a fixed number of neurons.

Regarding the activation function, the hyperbolic tangent was used in the neurons from the hidden layer and the identity function was used in both~~input and output layer's neurons.
The bias is absent in the input layer.
Naturally, the chromosome encodes the weights for each connection as well as the bias.

\subsection{Novelty Map and Novelty Map population}

Before describing a Novelty Map population, it is necessary to detail solely the Novelty Map.

\subsubsection{Novelty Map}

Novelty measures, when used as fitness functions, allow algorithms to keep track of what they have already seen, i.e., they provide the stepping stones to reach the objective \cite{reehuis2013novelty},\cite{lehman2011abandoning}.
Moreover, novelty can also be used to divide the space into points of interest, where each point is substantially different from each other.

Table~\ref{nmap_alg} describes the algorithm.
Basically, Novelty Map is a table with the most novel individuals according to a novelty metric. 
When a new input is presented to the map, a competition takes place where the cell with the closest weight array wins.
This winner cell is activated and can be used in many ways depending on the application (the novelty map population presents one way of using it).
Afterwards, the table is updated by substituting the weight array of the least novel cell (according to the novelty measure) with the input array if and only if the input array has higher novelty.
This way the table is always kept up to date.

\begin{table}[!h]
 \centering
\caption{Novelty Map Algorithm} 
\begin{tabular}{p{8cm}}
\hline
Parameters:
\begin{enumerate}
	\item $Max_n$: maximum size of the map;
	\item Novelty metric;
\end{enumerate}
Set the size of the map $n$ to zero\\
Infinite Loop:
\begin{enumerate}
\item When an input is presented to the novelty map do:
\item If the size of the map $n$ is smaller than $Max_n$
\begin{enumerate}
\item Insert the input in the map
\item Increment the size of the novelty map
\end{enumerate}
else
\begin{enumerate}
\item Evaluate the input's novelty with the novelty metric
\item If the input's novelty is higher than the lowest novelty from the samples inside the map
\begin{enumerate}
	\item Insert the input and remove the sample with the lowest novelty from the map
\end{enumerate}
\end{enumerate}
\item Return the weight array of the cell which is closest to the input
\end{enumerate} \\
\hline
\end{tabular}
\label{nmap_alg}
\end{table}

This table may share some similarities with the self-organizing map \cite{kohonen2001self} or even the neural gas \cite{fritzke1995growing}, but some important differences must be highlighted:
\begin{itemize}
	\item Independence on Input Frequency - Both neural gas and self-organizing map (SOM) are sensitive to the frequency of the input, forgetting previous experience if the input starts to concentrate on a small portion of the spectrum. The novelty map does not have this disadvantage, always retaining even the most rare occurrences if they are novel enough.
\item Cell's Efficiency - By ignoring the input frequency, fewer cells can be used to map the input space.  	
\end{itemize}

The novelty metric used in this article is the uniqueness.
Let $S$ be a set of arrays. 
The uniqueness is defined for an array $a_i$ in relation to the other arrays in $S$ with the following equation:
\begin{eqnarray}
	&U = S \setminus \{a_i\} \\
	&uniqueness = min_{a_k \in U}(dist(a_i,a_k)).
\end{eqnarray}
In other words, uniqueness of an array is the smallest distance  to the respective array for any array present in the set, excluding the array itself.
This novelty metric was chosen because of its simplicity and quality though any other novelty measure could have been used instead.

\subsubsection{Novelty Map Population}

The Novelty Map population is very similar to the SOM population \cite{vargas2013self,vargas2013aself}.
The only difference is the exchange of the SOM dynamics to the Novelty Map one.
As before, in addition to the cell's original weight array, subpopulations (see Section~\ref{subpopulation}) are present in all cells of the Novelty Map.
The original dynamics of the Novelty Map happens when a new input is presented, i.e., the cell which is closest to the input wins and the Novelty Map is updated.
Moreover, the winner cell and its subpopulation is used for some algorithm specific procedure.
For example, in reinforcement learning problems, the winner cell's subpopulation have one of its individuals selected to act on the environment.

\section{NOTC's Behavior}

Before going into the details of NOTC's behavior, it is necessary to explain two concepts (team and hall of fame concepts).

\subsection{Team}
\label{team_section}

\begin{figure}[!ht]
\centering
\includegraphics[scale=1.0]{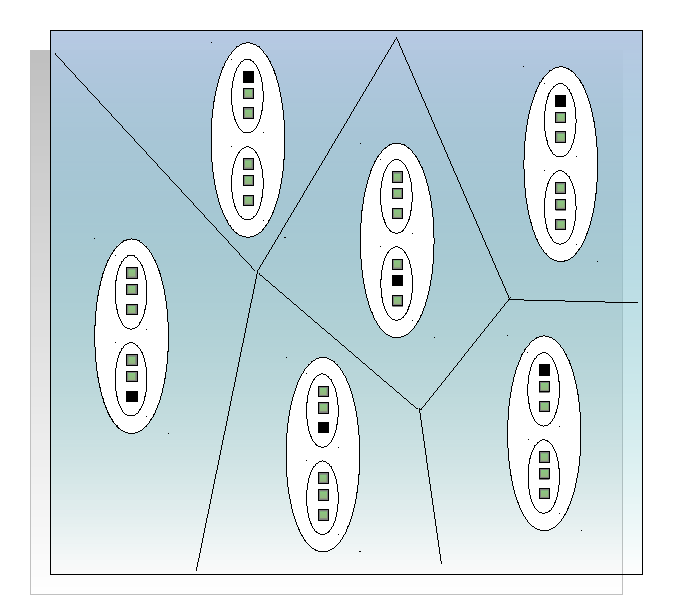}
\caption{Novelty Map Population with the black individuals (squares) forming a team. Once an individual from a certain cell is chosen to act in a trial, every time that cell is activated the same individual will be chosen to act. Suppose that all black squares are individuals which already acted in this trial. Therefore, they compose a team that is going to stay fixed until the end of the trial.}
\label{team}
\end{figure}

\begin{figure*}[!ht]
\centering
\includegraphics[scale=0.45]{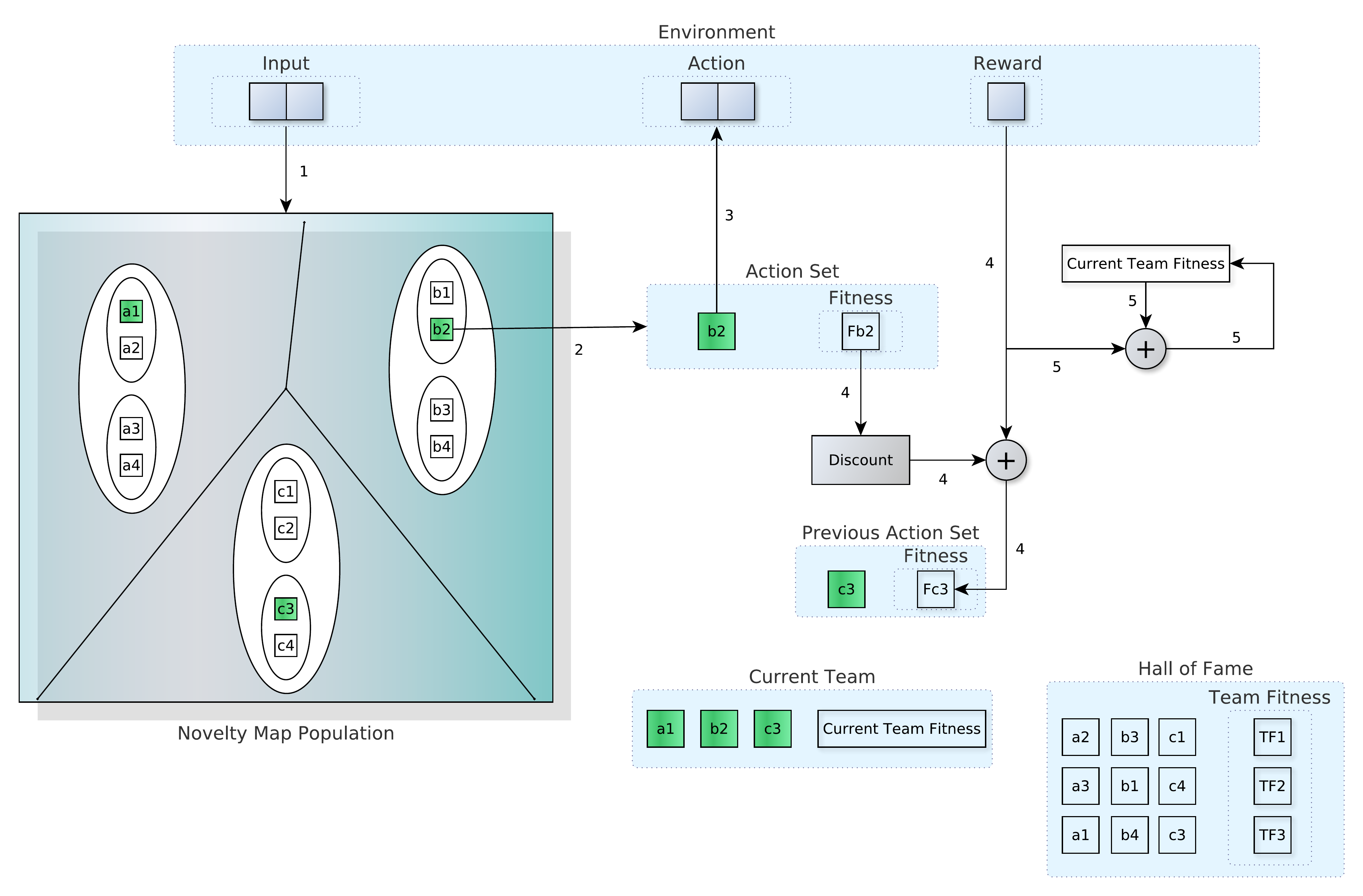}
\caption{NOTC's behavior.}
\label{behavior}
\end{figure*}

A reward is an evaluation of the last action and all of the actions that helped arrive in that last action.
One way of thinking about the problem is to have a set of individuals that are activated depending on the state, and let them receive the reward directly after its action, as well as a percentage of the fitness from the individual that acts in the next step.
Another way of thinking is to give the accumulated reward to all of the individuals that acted.
In fact, these two are the main reasoning behind Michigan and Pittsburgh LCS approaches with their pros and cons associated.
However they can be joined together if the team concept is used.

A team is a set of individuals each from a different cell in the Novelty Map Population (see Figure~\ref{team}).
When the cell is activated for the first time in the respective trial (i.e., the period from the start until the end of a run which is also called episode) an individual is chosen randomly.
Afterwards, the same previous individual is chosen every time this cell is activated in the same trial.
The team concepts is a consequence from this dynamic.
When a cell was not activated in a given trial, no individual is selected to be part of the team and therefore a don't care symbol is stored instead.

\subsection{Hall of Fame}

Without a place to store the best teams, this information would be lost and good combinations of individuals would be forgotten.
To prevent this, the hall of fame is created.
Hall of fame is the set of teams that received the highest accumulated reward.
Naturally, the accumulated reward is the sum of the rewards received by each individual in the given trial. 
In this article, the size of the hall of fame is fixed to half the number of best individuals in a cell.

\subsection{Behavior}

Figure~\ref{behavior} shows the NOTC's behavior.
This behavior is triggered when an input is received and happens throughout the trials.
The number showed in the arrows inside the figure correspond to a given step.
In the following these steps will be explained in detail:

\begin{enumerate}
	\item Novelty Map Population receives the input. Its cells compete for the input, with the winning cell having one of its individuals chosen to act (how one individual is chosen to act is explained in Section~\ref{team_section}).
	\item The chosen individual and its fitness compose the action set. 
	\item The chosen individual's neural network is activated, outputting the action to be performed.
	\item The individual that composed the previous action set has its fitness updated.
The fitness update is done using the Widrow-Hoff rule \cite{Widrow1960Adaptive}:
\begin{equation}
F = F + \eta(\hat{F} - F),
\end{equation}
where $\eta$ is the learning rate, $F$ is the current fitness and $\hat{F}$ is a new fitness estimate.
The fitness estimate of cell $cell$ and individual $c$ which were activated at time $t-1$ is given by the following equation:
\begin{equation}
	\hat{F}(c,cell)_{t-1} = R_{t-1} + \gamma \underset{c' \in cell'}{\operatorname{max}} \{F(c',cell')\},
\end{equation}
where $R$ is the reward received, $\gamma$ is the discount-factor and
$\underset{c \in cell}{\operatorname{max}} \{F(t)\}$ is the maximum fitness of individual $c'$ inside the activated cell $cell'$ at the current cycle $t$.
	\item Current team fitness accumulates the rewards received until the end of the trial. 
\end{enumerate}

There is though a single exception to how NOTC behaves.
After the evolution, the first trials are reserved for the teams in the hall of fame.
Therefore, each of the teams in the hall of fame have an trial where it must act and have its fitness updated.
This is important, otherwise a lucky team may stay for quite a long time as well as influence the evolution negatively.

\subsection{Evolution}

When $evolution\_trigger$ number of trials happened, the evolution is triggered. 
The following equation defines the $evolution\_trigger$:
\begin{equation}
	evolution\_trigger= S_{size}*\iota,
\end{equation}
where $S_{size}$ is the subpopulation size (best plus novel individuals) and $\iota$ is a parameter.

The evolution procedure consists of the following steps:
\begin{enumerate}
	\item For each cell, the first half of the best individuals is filled by the individuals present in the hall of fame teams and the second half with the fittest individuals according to their individual fitness. 
		Sometimes an individual is included multiple times, because it is part of both the fittest individuals as well as part of the hall of fame. 
		When a don't care symbol is present in the hall of fame team, a random individual from the cell is used.
	\item The remaining individuals are removed, resulting in an empty group of novel individuals.
	\item New novel individuals are created by using the differential evolution genetic operator (DE operator) or indexing with a chance of $50\%$ each.
		Therefore, for each novel individual a new individual is created with either one of the following:
\begin{itemize}
\item Indexing - A random individual from the population is copied;
\item DE operator - Consider that the number of best and novel individuals are the same. 
	The DE operator takes as base vector the best individual with the same index as the current novel individual to be created, in this way all best individuals will be used as base vectors of at least one novel individual. 
	To build the DE's mutant vector, three random individuals from the entire population (i.e., any individual from any subpopulation) are selected.
	The resulting trial vector is stored as the new novel individual.
\end{itemize}
\end{enumerate}


\section{Experiments' Settings}


In the following, experiments comparing NEAT with NOTC will be conducted in several variations of the mountain car problem.
All results are averaged over $30$ runs and only the best result among $100$ trials is plotted.

The NEAT code used is the 1.2.1 version of the NEAT C++ software package \cite{neatcode}.
Notice that although with different problem's settings (the initial position was randomized),  NEAT was previously applied to a discrete action Mountain Car \cite{whiteson2006evolutionary}.
Therefore, both the settings used in that paper and the settings present in the original package were evaluated.
In the end, the original package settings had better results, therefore the settings used for NEAT is the one provided with the software package (i.e. the same settings that was previously used in a double pole balancing task with success).
The parameters for NEAT are written in Table~\ref{para_neat}.

For NOTC, the settings are similar to the ones used in \cite{vargas2014novelty}.

\begin{table*}
\centering
\caption{Parameters for NEAT}
\begin{tabular}{ |c|l|c|l|c|l| }
	\hline
	Parameter & Value & Parameter & Value \\
	\hline
	\verb|trait_param_mut_prob| & $0.5$ & \verb|trait_mutation_power| & $1.0$ \\
	\verb|linktrait_mut_sig| & $1.0$ & \verb|nodetrait_mut_sig| & $0.5$ \\
	\verb|weigh_mut_power| & $2.5$ & \verb|recur_prob| & $0.00$ \\
	\verb|disjoint_coeff| & $1.0$ & \verb|excess_coeff| & $1.0$ \\
	\verb|mutdiff_coeff| & $0.4$ & \verb|compat_thresh| & $3.0$ \\
	\verb|age_significance| & $1.0$ & \verb|survival_thresh| & $0.20$ \\
	\verb|mutate_only_prob| & $0.25$ & \verb|mutate_random_trait_prob| & $0.1$ \\
	\verb|mutate_link_trait_prob| & $0.1$ &	\verb|mutate_node_trait_prob| & $0.1$ \\
	\verb|mutate_link_weights_prob| & $0.9$ & \verb|mutate_toggle_enable_prob| & $0.00$ \\
	\verb|mutate_gene_reenable_prob| & $0.000$ & \verb|mutate_add_node_prob| & $0.03$ \\ 
	\verb|mutate_add_link_prob| & $0.05$ & \verb|interspecies_mate_rate| & $0.001$ \\
	\verb|mate_multipoint_prob| & $0.6$ & \verb|mate_multipoint_avg_prob| & $0.4$ \\
	\verb|mate_singlepoint_prob| & $0.0$ & \verb|mate_only_prob| & $0.2$ \\
	\verb|recur_only_prob| & $0.0$ & \verb|pop_size| & $100$ \\
	\verb|dropoff_age| & $15$ & \verb|newlink_tries| & $20$ \\
	\verb|print_every| & $5$ & \verb|babies_stolen| & $0$ \\
	\verb|num_runs| & $1$ & &\\


	\hline
\end{tabular}
\label{para_neat}
\end{table*}

\begin{table*}
\centering
\caption{Parameters for Novelty-Organizing Team of Classifiers}
\begin{tabular}{ |c|l|l| }
	\hline
	& Parameter & Value \\ \hline
	\multirow{2}{*}{Differential Evolution} & CR & $0.2$\\
	   & F & random $\in [0.0, 2.0]$ \\ \hline
	\multirow{2}{*}{Novelty Map} & Number of Cells & $10$ \\
	 & Novelty Metric & Uniqueness \\
	\hline
	\multirow{5}{*}{Novelty-Organizing Team of Classifiers} & Widrow-hoff coefficient & $0.1$ \\ 
	 & Number of best individuals & $10$ \\
	 & Number of novel individuals & $10$ \\
	 & $\iota$ & $10$ \\
	 & Discount factor & $0.99$ \\
	 & Initial fitness for novel individuals & $-1$ \\
	 & Initial fitness for best individuals & $0$ \\
	 & Number of hidden nodes & $10$ \\
	\hline
\end{tabular}
\label{para_notc}
\end{table*}

\section{Experiment 1 - Continuous Action Mountain Car}

\begin{figure}[!ht]
\centering
\includegraphics[scale=0.3]{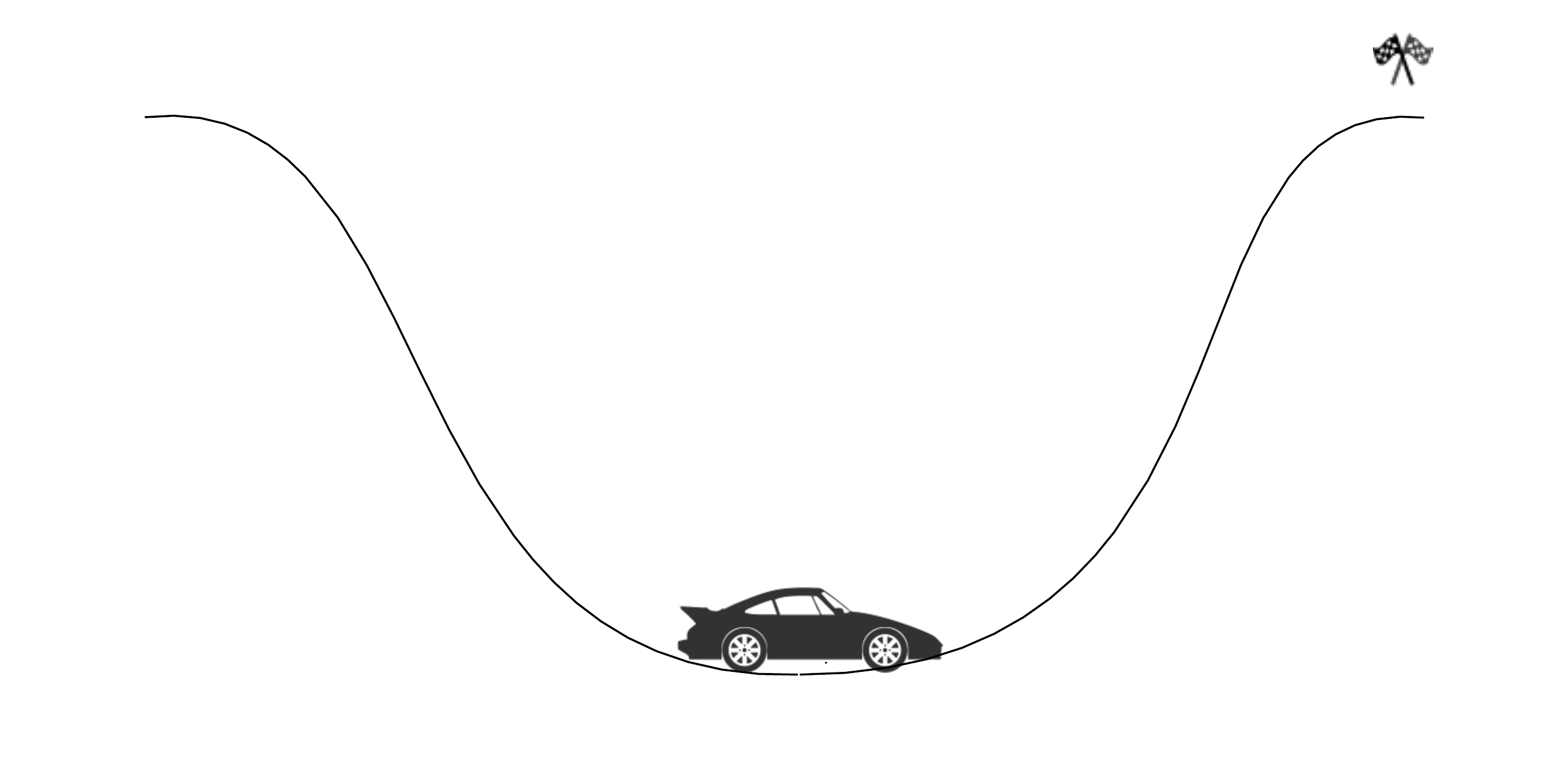}
\caption{Mountain car problem. The car's objective is to reach the flags uphill, although its acceleration is not enough to climb the mountain.}
\label{mountain_car}
\end{figure}

The mountain car problem \cite{sutton1996generalization} is shown in Figure~\ref{mountain_car}.
It is defined by the following equation:
\begin{equation}
\begin{aligned}
	&pos \in (-1.2,0.6) \\
	&v \in (-0.07,0.07) \\
	&a \in (-1,1) \\
	&v_{t+1} = v_{t} + (a_{t}) *0.001+\cos(3*pos_t)*(-0.0025) \\
	&pos_{t+1} = pos_t + v_{t+1},
\end{aligned}
\end{equation}
where $pos$ is the position of the car, $a$ is the car's action and $v$ is the velocity of the car.
The starting velocity and position are respectively $0.0$ and $-0.5$.
If $v < 0$ and $pos \le -1.2$, the velocity is set to zero.
When the car reaches $pos \ge 0.6$ the trial is terminated and the algorithm receives $0$ as reward.
In all other positions the algorithm receive $-1$ as a reward.
Moreover, if the algorithm's steps exceeds $10^3$ the trial is terminated and the common reward of $-1$ is returned to the algorithm.

\begin{figure}[!ht]
\centering
\includegraphics[scale=0.5]{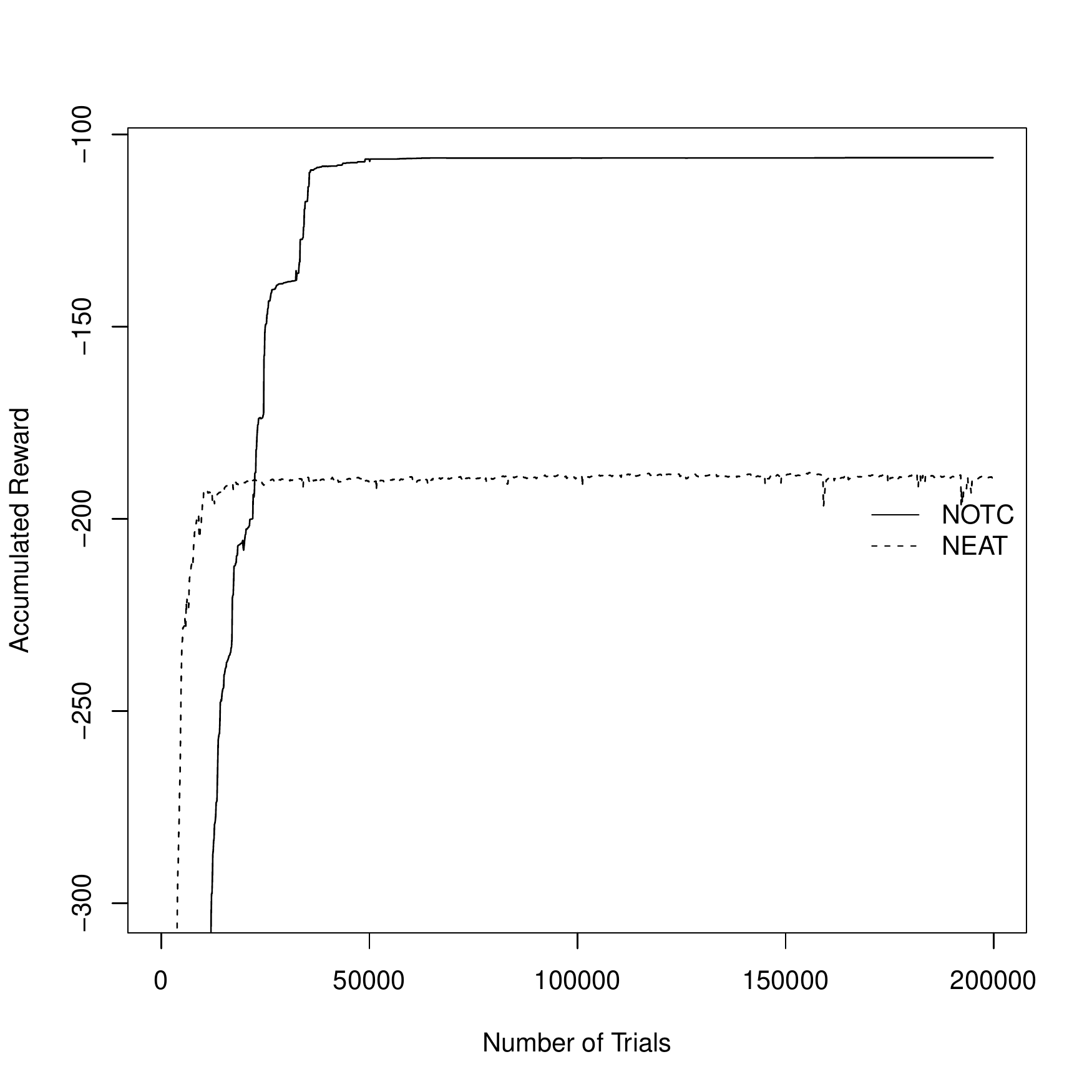}
\caption{Comparison of NOTC and NEAT in the continuous action mountain car problem.}
\label{continuous_mc}
\end{figure}

\begin{figure}[!ht]
\centering
\includegraphics[scale=0.5]{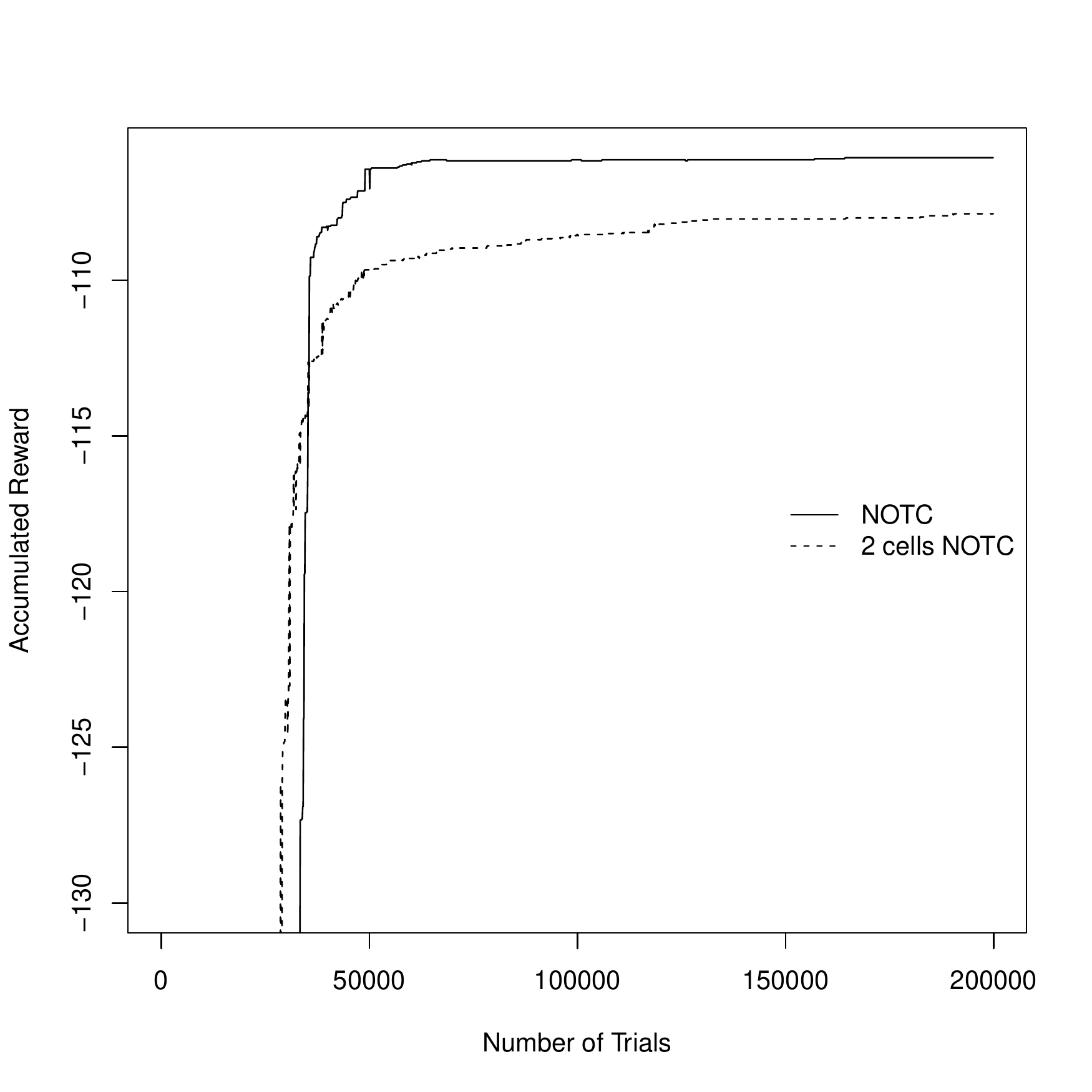}
\caption{Comparison of the original NOTC with a NOTC using a Novelty Map with only two cells in the continuous action mountain car problem.}
\label{continuous_mc2}
\end{figure}

NOTC takes more trials to converge, but surpasses NEAT in performance (see Figure~\ref{continuous_mc}).
The reasoning behind the better performance for NOTC lies in its ability of dividing the input space, creating smaller problem pieces that are easier to solve.
In fact, the importance of dividing the input space is verified by comparing with a NOTC with only two cells in the Novelty Map (see Figure~\ref{continuous_mc2}).
This NOTC is called two cells NOTC.
To make a fair comparison, the two cells NOTC has the same total number of individuals as the NOTC (i.e., both has the same initial diversity).
Therefore, the number of best and novel individuals in the two cells NOTC was increased to $50$.

For all experiments described in this paper, similar results were observed for their discrete action (-1,0 and 1) versions. Although the difference in the final performance between NEAT and NOTC was smaller.

\section{Experiment 2 - Continuous Action Mountain Car with Noise}

In the real world, sensors are always affected by noise.
To reflect this, a Gaussian noise is added to both the mountain car position and velocity every time they are read by the agent.
The added Gaussian noise is respectively $\mu=0$, $\sigma=0.06$ and $\mu=0$, $\sigma=0.009$ for the position and velocity of the car, where $\mu$ stands for the mean and $\sigma$ is the standard deviation.

\begin{figure}[!ht]
\centering
\includegraphics[scale=0.5]{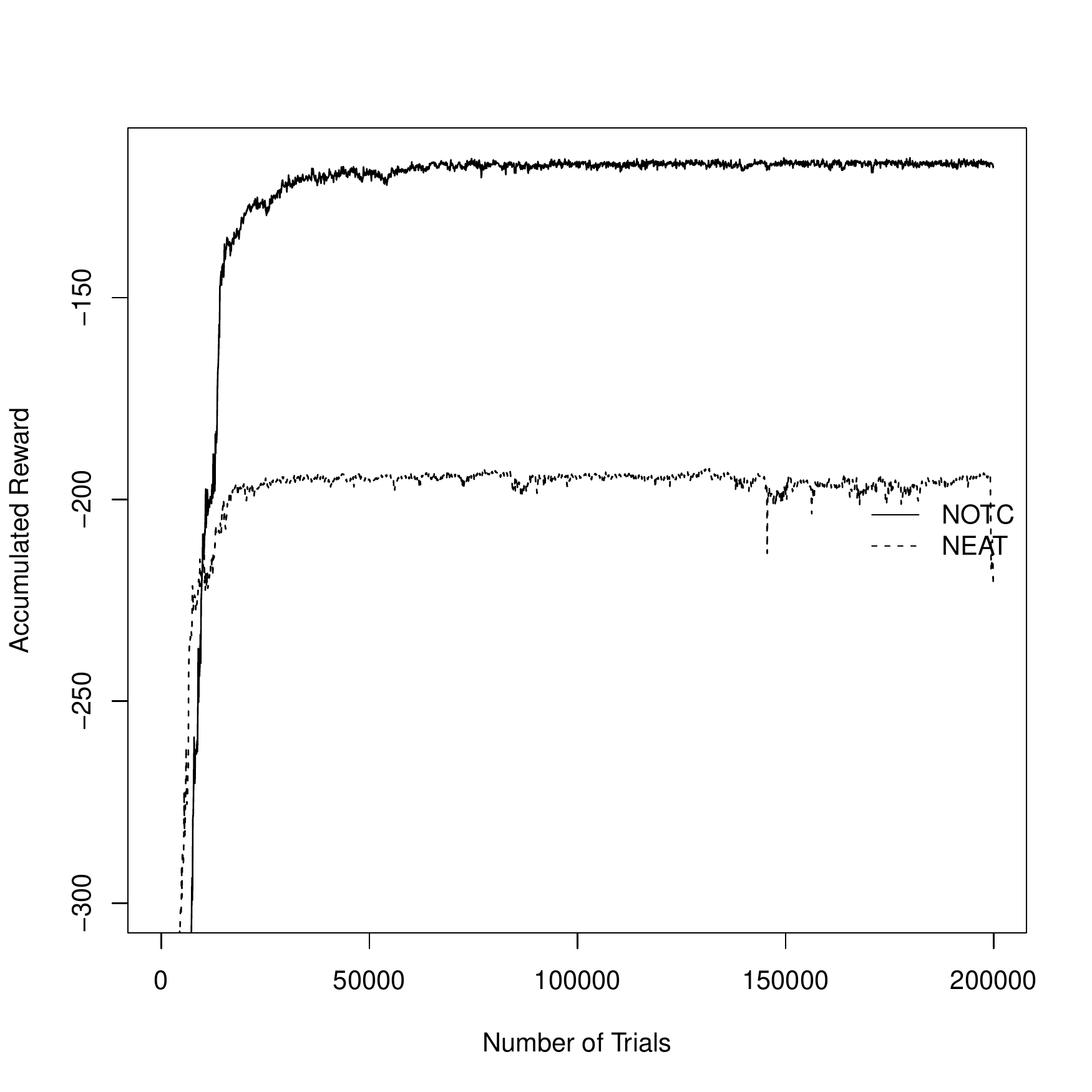}
\caption{Results of NOTC and NEAT when run on the noisy mountain car problem.}
\label{noisy_mc}
\end{figure}

Figure~\ref{noisy_mc} shows the results.
This result is very similar to the result from experiment 1, therefore the same observations made in experiment 1 can be made here.

\section{Experiment 3 - Continuous Action Mountain Car with Unstable Weather}

When driving a car, it is common for the weather to change from clear weather to rainy weather.
In that moment, for safety purposes, the maximum velocity decreases.
To reflect this, every $10000$ trials the maximum velocity changes from the original to the $(-0.04,0.04)$ range and vice-versa.
The motivation behind this problem is to verify the capability of an algorithm to adapt to changes in the environment.

\begin{figure}[!ht]
\centering
\includegraphics[scale=0.5]{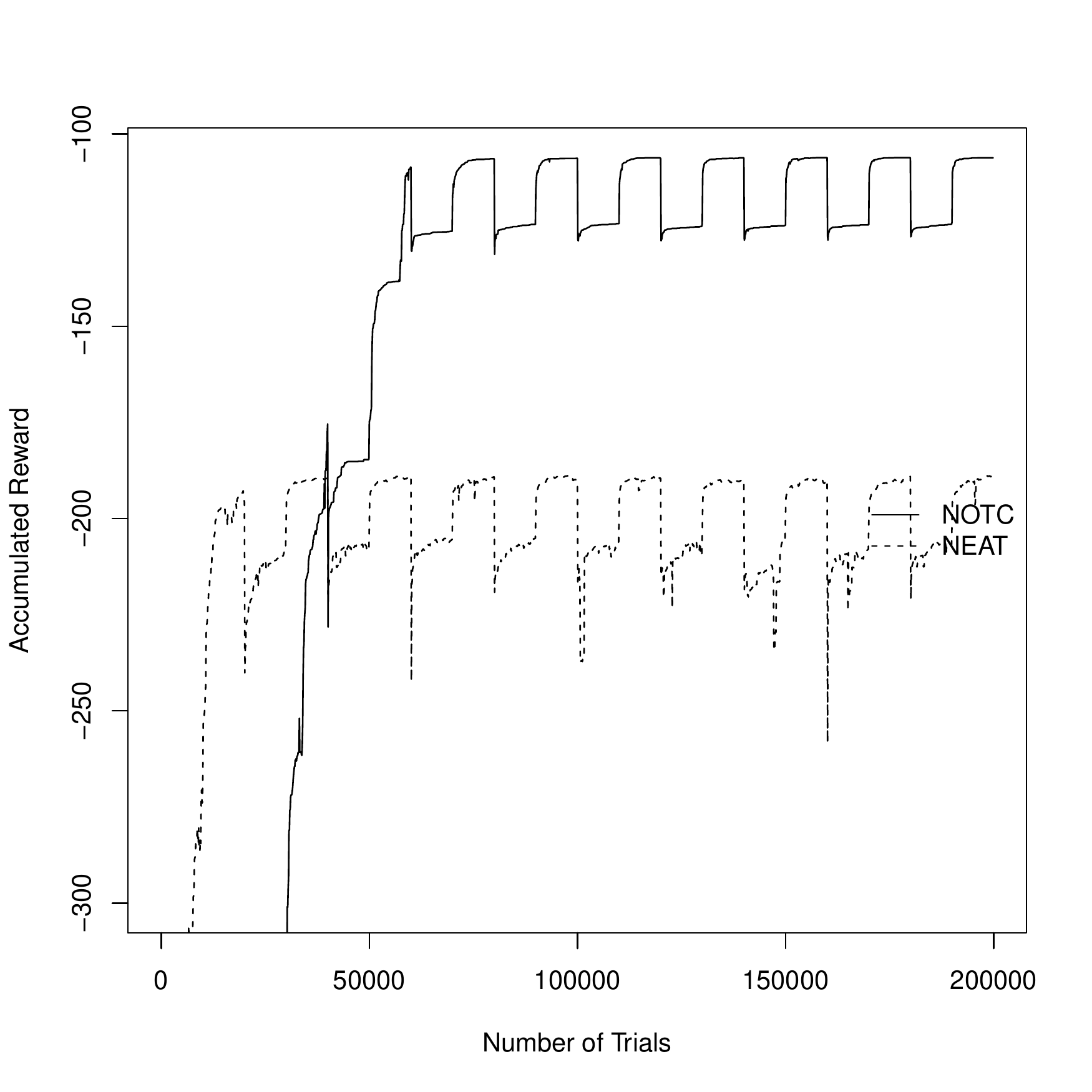}
\caption{NOTC and NEAT are compared in the unstable weather mountain car.}
\label{adaptive_mc}
\end{figure}

The results shown in Figure~\ref{adaptive_mc} demonstrate that NOTC achieves a better performance in both problems (reduced velocity or not) when compared with NEAT.
Moreover, NOTC has less variation of performance when the problem changes, which reveals that the solution found can easily change between both problems.
NEAT, on the other hand, has a very abrupt curve when the problem changes.

\section{Novelty Map Analysis}

Previous sections showed the behavior of the algorithm as a whole, but what can we say about the Novelty Map?
How fast does it self-organizes? 
Is it always changing?

\begin{figure}[!ht]
\centering
\includegraphics[scale=0.35]{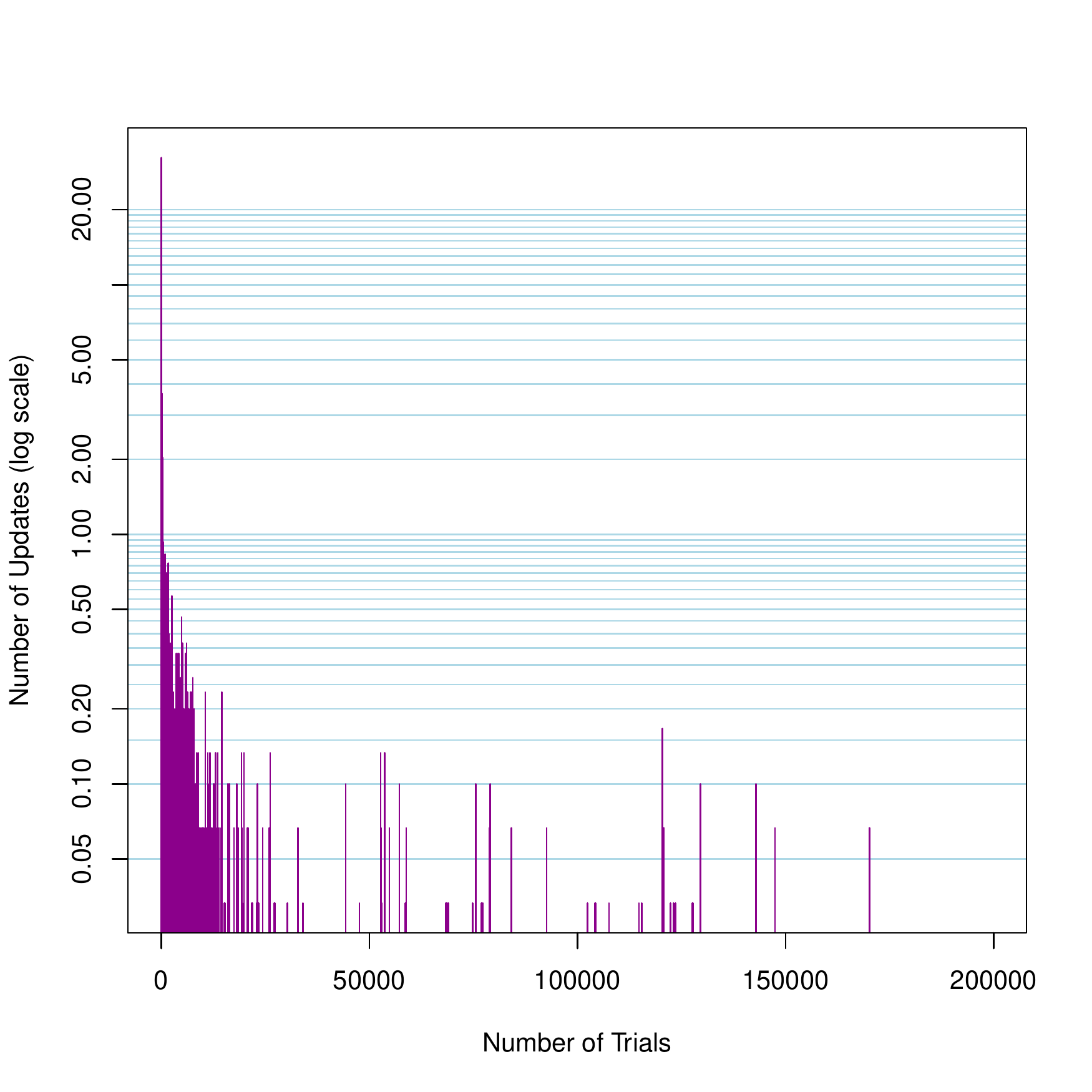}
\includegraphics[scale=0.35]{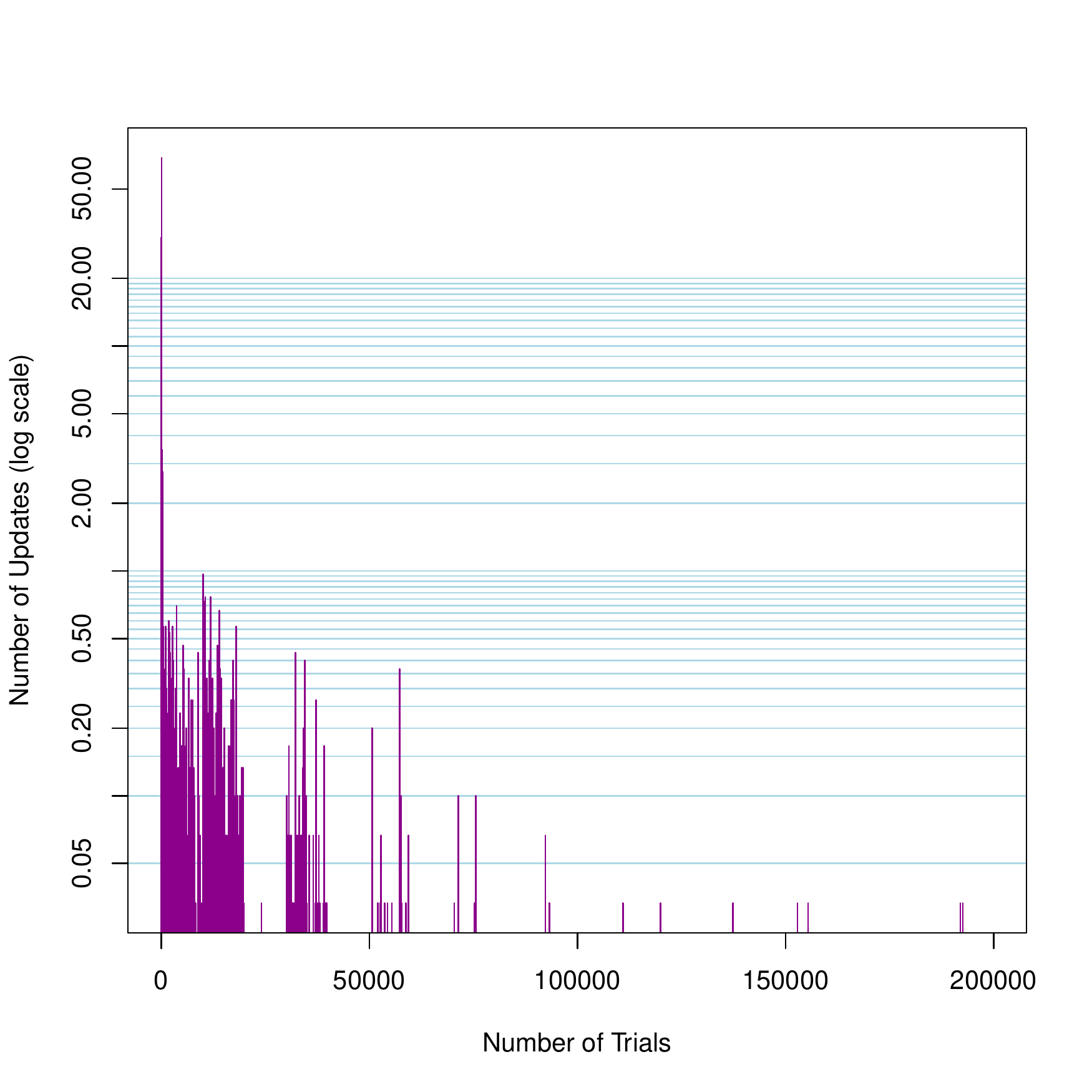}
\caption{Number of updates (changes in the value of cells) in the Novelty Map for the noisy mountain car (top) and unstable weather mountain car (below). The number of updates is in log scale.}
\label{nm_analysis}
\end{figure}

These questions can be answered by observing the number of times the value of cells are modified (updated) inside the Novelty Map (see Figure~\ref{nm_analysis}).
The number of updates decreases with the number of trials faster than exponentially.
Moreover, once the updates stop, the probability of another update appearing is very low.
In other words, the division of the input space is fixed after some time.
This fact allows for the evolution to focus on each of the smaller problems created.
The additional time required for the Novelty Map to stop updating also explains partially the reason why NOTC needs more trials than NEAT.
Naturally, this time also depends on the agent's exploration of the problem, the faster an agent explores the environment, the faster Novelty Map stops updating.

%
\IEEEpeerreviewmaketitle

\section{Conclusion}

In this article, NOTC was described in detail.
Moreover, NOTC and NEAT were compared in a continuous action mountain car and two variations of it, one with noise and the other with the problem dynamics changing throughout the experiments.
The experiments revealed a trade-off between the algorithms.
NOTC achieved better performance in all of the problems, although NEAT needed less trials to converge.
That is, despite the fact that NEAT evolves both the topology and parameters of the neural network, allowing for more robust and complex models, it was surpassed in performance by NOTC using a divide and conquer approach with a fixed topology neural network.
The division of the input space was shown to be the reason why NOTC has a better performance.
NOTC, however, takes more time to converge due to the additional time involved in learning the division of the problem.

Thus, the verified trade-off should be to some extent present when comparing a divide and conquer with a complexification approach.
Having said that, it should be noticed that both algorithms can improve, alleviating the trade-off.
In special, the divide and conquer strategy was very far from its full potential, since the division was done over a space where the problem is non-separable (i.e. the input space for the mountain car was not enough to define a state space).


\section*{Acknowledgment}

This work was supported in part by JSPS KAKENHI Grant Number 24560499.



\bibliographystyle{IEEEtran}
\bibliography{sigproc}

\begin{thebibliography}{10}
\providecommand{\url}[1]{#1}
\csname url@samestyle\endcsname
\providecommand{\newblock}{\relax}
\providecommand{\bibinfo}[2]{#2}
\providecommand{\BIBentrySTDinterwordspacing}{\spaceskip=0pt\relax}
\providecommand{\BIBentryALTinterwordstretchfactor}{4}
\providecommand{\BIBentryALTinterwordspacing}{\spaceskip=\fontdimen2\font plus
\BIBentryALTinterwordstretchfactor\fontdimen3\font minus
  \fontdimen4\font\relax}
\providecommand{\BIBforeignlanguage}[2]{{%
\expandafter\ifx\csname l@#1\endcsname\relax
\typeout{** WARNING: IEEEtran.bst: No hyphenation pattern has been}%
\typeout{** loaded for the language `#1'. Using the pattern for}%
\typeout{** the default language instead.}%
\else
\language=\csname l@#1\endcsname
\fi
#2}}
\providecommand{\BIBdecl}{\relax}
\BIBdecl

\bibitem{stanley2002evolving}
K.~Stanley and R.~Miikkulainen, ``Evolving neural networks through augmenting
  topologies,'' \emph{Evolutionary computation}, vol.~10, no.~2, pp. 99--127,
  2002.

\bibitem{vargas2014novelty}
D.~V. Vargas, H.~Takano, and J.~Murata, ``Novelty-organizing team of
  classifiers-a team-individual multi-objective approach to reinforcement
  learning,'' in \emph{SICE Annual Conference (SICE), 2014 Proceedings of
  the}.\hskip 1em plus 0.5em minus 0.4em\relax IEEE, 2014, pp. 1785--1792.

\bibitem{holland1977cognitive}
J.~H. Holland and J.~S. Reitman, ``Cognitive systems based on adaptive
  algorithms,'' \emph{ACM SIGART Bulletin}, no.~63, pp. 49--49, 1977.

\bibitem{holmes2002learning}
J.~H. Holmes, P.~L. Lanzi, W.~Stolzmann, and S.~W. Wilson, ``Learning
  classifier systems: New models, successful applications,'' \emph{Information
  Processing Letters}, vol.~82, no.~1, pp. 23--30, 2002.

\bibitem{floreano2008neuroevolution}
D.~Floreano, P.~D{\"u}rr, and C.~Mattiussi, ``Neuroevolution: from
  architectures to learning,'' \emph{Evolutionary Intelligence}, vol.~1, no.~1,
  pp. 47--62, 2008.

\bibitem{urbanowicz2009learning}
R.~Urbanowicz and J.~Moore, ``Learning classifier systems: a complete
  introduction, review, and roadmap,'' \emph{Journal of Artificial Evolution
  and Applications}, vol. 2009, p.~1, 2009.

\bibitem{lanzi2000roadmap}
P.~Lanzi and R.~Riolo, ``A roadmap to the last decade of learning classifier
  system research (from 1989 to 1999),'' \emph{Learning Classifier Systems},
  pp. 33--61, 2000.

\bibitem{wilson2002classifiers}
S.~W. Wilson, ``Classifiers that approximate functions,'' \emph{Natural
  Computing}, vol.~1, no. 2-3, pp. 211--234, 2002.

\bibitem{butz2008function}
M.~Butz, P.~Lanzi, and S.~Wilson, ``Function approximation with {XCS}:
  Hyperellipsoidal conditions, recursive least squares, and compaction,''
  \emph{Evolutionary Computation, IEEE Transactions on}, vol.~12, no.~3, pp.
  355--376, 2008.

\bibitem{tran2007xcsf}
H.~Tran, C.~Sanza, Y.~Duthen, and T.~Nguyen, ``{XCSF} with computed continuous
  action,'' in \emph{Genetic And Evolutionary Computation Conference:
  Proceedings of the 9 th annual conference on Genetic and evolutionary
  computation}, vol.~7, no.~11, 2007, pp. 1861--1869.

\bibitem{valenzuela1991fuzzy}
M.~Valenzuela-Rend{\'o}n, ``The fuzzy classifier system: A classifier system
  for continuously varying variables,'' in \emph{Proceedings of the Fourth
  International Conference on Genetic Algorithms pp346-353, Morgan Kaufmann I},
  vol. 991, 1991, pp. 223--230.

\bibitem{bull2002accuracy}
L.~Bull and T.~O'Hara, ``Accuracy-based neuro and neuro-fuzzy classifier
  systems,'' in \emph{Proceedings of the Genetic and Evolutionary Computation
  Conference}.\hskip 1em plus 0.5em minus 0.4em\relax Morgan Kaufmann
  Publishers Inc., 2002, pp. 905--911.

\bibitem{casillas2007fuzzy}
J.~Casillas, B.~Carse, and L.~Bull, ``{Fuzzy-XCS}: A michigan genetic fuzzy
  system,'' \emph{Fuzzy Systems, IEEE Transactions on}, vol.~15, no.~4, pp.
  536--550, 2007.

\bibitem{bull2002using}
L.~Bull, ``On using constructivism in neural classifier systems,''
  \emph{Parallel problem solving from nature-PPSN VII}, pp. 558--567, 2002.

\bibitem{iqbal2012xcsr}
M.~Iqbal, W.~N. Browne, and M.~Zhang, ``Xcsr with computed continuous action,''
  in \emph{AI 2012: Advances in Artificial Intelligence}.\hskip 1em plus 0.5em
  minus 0.4em\relax Springer, 2012, pp. 350--361.

\bibitem{lanzi2005xcs}
P.~Lanzi, D.~Loiacono, S.~Wilson, and D.~Goldberg, ``{XCS} with computed
  prediction in multistep environments,'' in \emph{Proceedings of the 2005
  conference on Genetic and evolutionary computation}.\hskip 1em plus 0.5em
  minus 0.4em\relax ACM, 2005, pp. 1859--1866.

\bibitem{Twardowski1993a}
K.~Twardowski, ``{Credit Assignment for Pole Balancing with Learning Classifier
  Systems},'' pp. 238--245.

\bibitem{bonarini1996evolutionary}
A.~Bonarini, ``Evolutionary learning of fuzzy rules: competition and
  cooperation,'' in \emph{Fuzzy Modelling}.\hskip 1em plus 0.5em minus
  0.4em\relax Springer, 1996, pp. 265--283.

\bibitem{lanzi2006classifier}
P.~L. Lanzi, D.~Loiacono, S.~W. Wilson, and D.~E. Goldberg, ``Classifier
  prediction based on tile coding,'' in \emph{Proceedings of the 8th annual
  conference on Genetic and evolutionary computation}.\hskip 1em plus 0.5em
  minus 0.4em\relax ACM, 2006, pp. 1497--1504.

\bibitem{stalph2012learning}
P.~Stalph and M.~Butz, ``Learning local linear jacobians for flexible and
  adaptive robot arm control,'' \emph{Genetic programming and evolvable
  machines}, vol.~13, no.~2, pp. 137--157, 2012.

\bibitem{butz2008context}
M.~V. Butz and O.~Herbort, ``Context-dependent predictions and cognitive arm
  control with xcsf,'' in \emph{Proceedings of the 10th annual conference on
  Genetic and evolutionary computation}.\hskip 1em plus 0.5em minus 0.4em\relax
  ACM, 2008, pp. 1357--1364.

\bibitem{bonarini2000fuzzy}
A.~Bonarini, C.~Bonacina, and M.~Matteucci, ``Fuzzy and crisp representations
  of real-valued input for learning classifier systems,'' \emph{Learning
  Classifier Systems}, pp. 107--124, 2000.

\bibitem{howard2009towards}
G.~Howard, L.~Bull, and P.~Lanzi, ``Towards continuous actions in continuous
  space and time using self-adaptive constructivism in neural {XCSF},'' in
  \emph{Proceedings of the 11th Annual conference on Genetic and evolutionary
  computation}.\hskip 1em plus 0.5em minus 0.4em\relax ACM, 2009, pp.
  1219--1226.

\bibitem{vargas2013self}
D.~V. Vargas, H.~Takano, and J.~Murata, ``Self organizing classifiers: first
  steps in structured evolutionary machine learning,'' \emph{Evolutionary
  Intelligence}, vol.~6, no.~2, pp. 57--72, 2013.

\bibitem{vargas2013aself}
------, ``Self organizing classifiers and niched fitness,'' in
  \emph{Proceedings of the fifteenth annual conference on Genetic and
  evolutionary computation conference}.\hskip 1em plus 0.5em minus 0.4em\relax
  ACM, 2013, pp. 1109--1116.

\bibitem{vargas2013continuous}
------, ``Continuous adaptive reinforcement learning with the evolution of self
  organizing classifiers,'' in \emph{Development and Learning and Epigenetic
  Robotics (ICDL), 2013 IEEE Third Joint International Conference on}.\hskip
  1em plus 0.5em minus 0.4em\relax IEEE, 2013, pp. 1--2.

\bibitem{vargas2014noc}
------, ``Novelty-organizing classifiers applied to classification and
  reinforcement learning: towards flexible algorithms,'' in \emph{Genetic and
  Evolutionary Computation Conference, {GECCO} '14, Vancouver, BC, Canada, July
  12-16, 2014, Companion Material Proceedings}, 2014, pp. 81--82.

\bibitem{angeline1994evolutionary}
P.~J. Angeline, G.~M. Saunders, and J.~B. Pollack, ``An evolutionary algorithm
  that constructs recurrent neural networks,'' \emph{Neural Networks, IEEE
  Transactions on}, vol.~5, no.~1, pp. 54--65, 1994.

\bibitem{kassahun2005efficient}
Y.~Kassahun and G.~Sommer, ``Efficient reinforcement learning through
  evolutionary acquisition of neural topologies,'' in \emph{In 13th European
  Symposium on Artificial Neural Networks (ESANN)}.\hskip 1em plus 0.5em minus
  0.4em\relax Citeseer, 2005.

\bibitem{yao1997new}
X.~Yao and Y.~Liu, ``A new evolutionary system for evolving artificial neural
  networks,'' \emph{Neural Networks, IEEE Transactions on}, vol.~8, no.~3, pp.
  694--713, 1997.

\bibitem{reehuis2013novelty}
E.~Reehuis, M.~Olhofer, M.~Emmerich, B.~Sendhoff, and T.~B{\"a}ck, ``Novelty
  and interestingness measures for design-space exploration,'' in
  \emph{Proceeding of the fifteenth annual conference on Genetic and
  evolutionary computation conference}.\hskip 1em plus 0.5em minus 0.4em\relax
  ACM, 2013, pp. 1541--1548.

\bibitem{lehman2011abandoning}
J.~Lehman and K.~O. Stanley, ``Abandoning objectives: Evolution through the
  search for novelty alone,'' \emph{Evolutionary computation}, vol.~19, no.~2,
  pp. 189--223, 2011.

\bibitem{kohonen2001self}
T.~Kohonen, \emph{Self-organizing maps}.\hskip 1em plus 0.5em minus 0.4em\relax
  Springer, 2001, vol.~30.

\bibitem{fritzke1995growing}
B.~Fritzke \emph{et~al.}, ``A growing neural gas network learns topologies,''
  \emph{Advances in neural information processing systems}, vol.~7, pp.
  625--632, 1995.

\bibitem{Widrow1960Adaptive}
\BIBentryALTinterwordspacing
B.~Widrow and M.~E. Hoff, ``{Adaptive Switching Circuits},'' in \emph{1960
  {IRE} {WESCON} Convention Record, Part 4}.\hskip 1em plus 0.5em minus
  0.4em\relax New York: {IRE}, 1960, pp. 96--104. [Online]. Available:
  \url{http://isl-www.stanford.edu/\~{}widrow/papers/c1960adaptiveswitching.pdf}
\BIBentrySTDinterwordspacing

\bibitem{neatcode}
K.~Stanley, I.~Karpov, B.~Erkin, and D.~Thomas, ``{NEAT C++},''
  \url{http://nn.cs.utexas.edu/keyword?neat-c}, 2001--2011.

\bibitem{whiteson2006evolutionary}
S.~Whiteson and P.~Stone, ``Evolutionary function approximation for
  reinforcement learning,'' \emph{The Journal of Machine Learning Research},
  vol.~7, pp. 877--917, 2006.

\bibitem{sutton1996generalization}
R.~S. Sutton, ``Generalization in reinforcement learning: Successful examples
  using sparse coarse coding,'' in \emph{Advances in Neural Information
  Processing Systems 8}, 1996.

\end{thebibliography}
%


\end{document}